\documentclass[
]{ceurart}

\usepackage{url}
\usepackage{amsmath}
\usepackage{amsfonts}
\usepackage{graphicx}
\usepackage{hyperref}
\usepackage{changepage}
\usepackage{bbm}
\usepackage{tabularx}
\usepackage{booktabs}
\usepackage{multirow}
\usepackage{tikz}
\usepackage{booktabs}
\usepackage{array}
\usepackage{makecell}
\usepackage{multirow}

\begin{document}



\title{A Comparative Analysis of Counterfactual Explanation Methods for Text Classifiers}

\author[1]{Stephen McAleese}[
    email=stephen.mcaleese@ucdconnect.ie
]

\author[1,2]{Mark Keane}

\address[1]{School of Computer Science, University College Dublin, Dublin, Ireland}

\address[2]{Insight Centre for Data Analytics, Dublin, Ireland}

\begin{abstract}
Counterfactual explanations can be used to interpret and debug text classifiers by producing minimally altered text inputs that 
change a classifier's output. In this work, we evaluate five methods for generating counterfactual explanations for a BERT text classifier on two datasets using three evaluation metrics. The results of our experiments suggest that established white-box substitution-based methods are effective at generating valid counterfactuals that change the classifier's output. In contrast, newer methods based on large language models (LLMs) excel at producing natural and linguistically plausible text counterfactuals but often fail to generate valid counterfactuals that alter the classifier's output. Based on these results, we recommend developing new counterfactual explanation methods that combine the strengths of established gradient-based approaches and newer LLM-based techniques to generate high-quality, valid, and plausible text counterfactual explanations.
\end{abstract}


\maketitle

\section{Introduction}

The rapid advancement of machine learning (ML) and deep learning (DL) models has led to their widespread application in real-world tasks. Due to algorithmic improvements, increased computational power, and dataset sizes, the performance of models powered by deep neural networks has surpassed traditional ML algorithms such as logistic regression and support vector machines (SVMs) on many tasks such as speech recognition, language translation, summarization, and sentiment classification \cite{lecunDeepLearning2015,sevillaComputeTrendsThree2022}. Today, many high-performing NLP models such as BERT \cite{devlin2019bertpretrainingdeepbidirectional} and GPT-4 \cite{openai2024gpt4technicalreport} are large language models (LLMs) consisting of dozens of transformer layers and millions or billions of parameters.

While these models have achieved remarkable performance across a wide range of tasks, their complexity and scale decreases their transparency and explainability. Consequently, many leading NLP models are considered to be "black boxes", providing outputs to users without explaining how or why they were produced \cite{doranWhatDoesExplainable2017a,adadiPeekingBlackBoxSurvey2018}. Their lack of interpretability and explainability hinders their trustworthiness, fairness and the ability of model developers to identify potential flaws \cite{arrietaExplainableArtificialIntelligence2019,doshi-velezRigorousScienceInterpretable2017}. For example, a black-box model may contain hidden biases or rely on spurious features to make decisions. Furthermore, the lack of transparency of these models may impede their deployment in sensitive domains such as healthcare, finance, and legal applications \cite{adadiPeekingBlackBoxSurvey2018}.

\begin{figure}[ht]
    \centering
    \includegraphics[width=9cm]{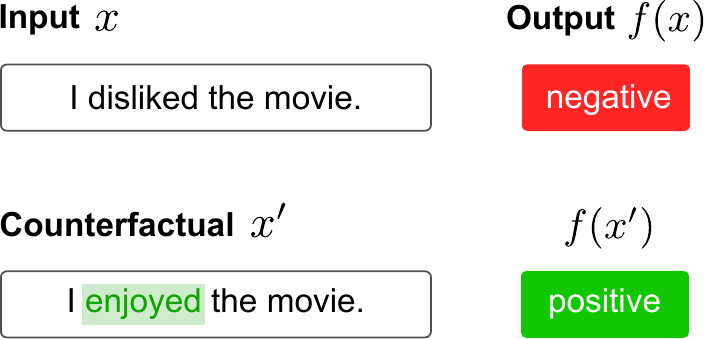}
    \caption{An example of a counterfactual explanation.}
    \label{fig:text-counterfactual-example}
\end{figure}

The problems associated with black-box models have motivated the development of the fields of interpretability and explainable AI (XAI) which aim to develop techniques to understand the inner workings of these models and generate explanations for model outputs \cite{gilpinExplainingExplanationsOverview2019,gunningDARPAExplainableArtificialIntelligenceProgram}. Among the various techniques introduced for explaining ML models, research activity on counterfactual explanations has increased significantly since 2019 due to their strong support from psychological research and other benefits \cite{keaneIfOnlyWe2021}.

Counterfactual explanations demonstrate how an input can be minimally altered to change a classifier's output in order to show which parts of the input are important for determining the model's output \cite{wachterCounterfactualExplanationsOpening2017a}. For example, given the sentence \textit{"I liked the movie"} classified as positive sentiment, a counterfactual input \textit{"I hated the movie"} classified as negative sentiment helps users understand the classifier's behavior by highlighting the influence of the verb on the model's output. Since their introduction, counterfactual explanations have been widely applied to traditional ML problems involving explicit features and tabular datasets \cite{keaneIfOnlyWe2021}. More recently, new methods have been developed to generate counterfactual explanations for deep learning models trained on image and text classification tasks \cite{kennyGeneratingPlausibleCounterfactual2020c}.

Several methods have been developed for generating counterfactual explanations for natural language processing (NLP) models such as text classifiers. However, many of the papers that introduced these methods rely on baselines from the adversarial robustness literature rather than the counterfactual explanations literature due to a lack of other methods to compare against. Furthermore, the lack of standardization of evaluation metrics and baselines in the literature increases the difficulty of evaluating different methods, making it challenging to compare their relative effectiveness or understand the strengths and weaknesses of each method.

To address these issues, we conduct a comparative study of five methods for generating counterfactual explanations for a text classifier across two datasets, aiming to identify the most effective method as measured by three standardized evaluation metrics. The text classifier we aim to explain is a BERT-base model \cite{devlin2019bertpretrainingdeepbidirectional} fine-tuned on each dataset although any other black-box model could be used since counterfactual explanations are model-agnostic.

Due to the effectiveness of large language models (LLMs) at producing plausible text outputs and handling other NLP tasks, many methods for generating counterfactual explanations for text classifiers involve LLMs. Early methods such as CLOSS (2021) \cite{fernTextCounterfactualsLatent2021a} use LLMs to generate plausible word substitutes. Other methods such as Polyjuice (2021) \cite{wuPolyjuiceGeneratingCounterfactuals2021a} use fine-tuned LLM models to generate counterfactuals. More recently, due to advancements in the performance of LLMs, methods such as FIZLE (2024) \cite{bhattacharjeeZeroshotLLMguidedCounterfactual2024} have been developed which use only general-purpose LLMs and task-specific system prompts to generate counterfactuals without the need for additional fine-tuning or task-specific architectures. These methods leverage the instruction-following and textual understanding capabilities of recent state-of-the-art LLMs such as GPT-4 \cite{openai2024gpt4technicalreport} to generate high-quality counterfactuals. While these recent methods greatly simplify the process of generating text counterfactuals, it remains unclear whether they outperform older, more specialized methods and how their strengths and weaknesses differ.

Given these uncertainties, our research aims to address two primary research questions: 1) Which methods for generating counterfactual explanations for text classifiers are most effective? and 2) Do newer, simpler methods that prompt general-purpose LLMs to generate counterfactuals outperform older, more specialized methods?

The following list describes the contributions of this paper:

\begin{enumerate}
    \item A review of the literature describing several past methods for generating counterfactual explanations for text classifiers.
    \item A set of three standardized evaluation metrics for measuring the effectiveness of methods for generating text counterfactual explanations. 
    \item A comparative study of five methods for generating text counterfactual explanations across two datasets using three evaluation metrics to compare the effectiveness of each method.
\end{enumerate}

The rest of this paper is organized as follows: Section 2 provides a background on model explanations and counterfactual explanations, Section 3 describes a literature review of past methods for generating counterfactual explanations including detailed explanations of each of the five methods selected for our comparative study, Section 4 outlines the experimental setup for our comparative study, Section 5 presents the results of our experiments, and Section 6 summarizes our conclusions and suggests directions for future research.

\section{Background}

\subsection{Model explanation}

Our research addresses the challenge of explaining the outputs from a binary classifier powered by a BERT-base neural language model \cite{devlin2019bertpretrainingdeepbidirectional}. Two broad approaches to explaining the predictions of deep neural networks are explaining the \textit{processing} of data or explaining the \textit{representation} of data inside the neural network \cite{gilpinExplainingExplanationsOverview2019}. The former approach seeks to answer the question, \textit{"Why does this particular input lead to that particular output?"} while the latter aims to answer the question \textit{"What information does the network contain?"}. Our work focuses on the first question, the problem of explaining why the model produced a particular output. While explanations of network representations such as feature visualizations \cite{olahFeatureVisualization2017} for individual neurons can offer valuable information, they provide limited insight into the reasons why a model produced a particular output. Therefore, we focus on explaining the processing of data to gain a more high-level understanding of model behavior.

Some popular methods for understanding ML classifiers include LIME \cite{ribeiroWhyShouldTrust2016a} and saliency maps \cite{simonyanDeepConvolutionalNetworks2014,smilkov2017smoothgradremovingnoiseadding,montavonMethodsForInterpretingDNNs}. LIME explains models by creating a simple, interpretable linear model that approximates the more complex model locally. Saliency maps highlight which components of the input have the most significant influence on the output typically by calculating the partial derivative of the output with respect to each input component. Although LIME has proven useful for explaining various classifiers, approximating the behavior of a language model with millions of parameters which accepts unstructured text as input is challenging. Saliency maps can also be helpful but provide non-contrastive explanations, which can limit the intuitiveness of their explanations for users \cite{jacovi2021contrastiveexplanationsmodelinterpretability,yin2022interpretinglanguagemodelscontrastive}.

To explain our black-box classifier, we instead turn to counterfactual explanations which provide model-agnostic, contrastive explanations which do not require a detailed understanding of the model's internal mechanisms \cite{stepinSurveyOfConstrastiveAndCounterfactualExplanationMethods}.

\subsection{Counterfactual explanations}

The concept of explanation is closely linked to causality and counterfactual events are a key feature of causal reasoning. An event C can be said to have caused E if, under some hypothetical counterfactual case the event C did not occur, E would not have occurred \cite{millerExplanationArtificialIntelligence2019}.

Research from psychology and the social sciences has found that people generally seek out contrastive explanations for events. This means that instead of asking why event $P$ occurred, people tend to seek explanations that explain why event $P$ happened rather than some alternative event $Q$ \cite{millerExplanationArtificialIntelligence2019}.  Furthermore, people are selective and tend to select one or two primary causes from many possible potential causes.

These findings have motivated the development of methods for generating counterfactual explanations which provide contrastive explanations for model outputs \cite{wachterCounterfactualExplanationsOpening2017a}. Counterfactual explanations are usually given in the form of a counterfactual instance or counterfactual: given an input $x$ and a classifier $f : x \rightarrow y$, a counterfactual is a minimally altered input $x'$ that produces a different outcome $y'$ from the classifier. The task of generating a counterfactual input $x'$ for a classifier can be formulated mathematically as the following optimization problem: 

\begin{equation}
argmin_{x'}d(x, x') \ subject \ to \ f(x') = y'
\end{equation}

For example, in the context of a sentiment classifier the sentence, \textit{"I loved the movie"} is classified as positive sentiment by the classifier. The counterfactual sentence, \textit{"I hated the movie"} is a minimally altered input classified as negative sentiment. In this case, it's clear that the verb "loved" or "hated" causes the output of the model.

The quality of generated counterfactuals can vary depending on several factors. Researchers in the field of explainable AI generally agreed that high-quality counterfactuals have the following properties \cite{vermaCounterfactualExplanationsMachinec}:

\begin{itemize}
    \item \textbf{Validity}: The counterfactual instance $x'$ should be classified as a member of a different class $y'$ to the original class $y$. This ensures that the counterfactual actually causes a change in the model's decision, and is informative for understanding the model's behavior.
    \item \textbf{Sparsity}: The number of input features changed to create the counterfactual should be minimized to highlight the small number of significant input features that contributed to the output. Sparse counterfactuals are typically more interpretable for end-users.
    \item \textbf{Plausibility:} The feature changes in the counterfactual should be realistic and fall within the distribution of inputs the model has been trained on and expects to encounter.
\end{itemize}

\section{Methods}

This section provides detailed explanations of several methods developed in recent years for generating counterfactual explanations for text classifiers. These methods are also included in our comparative study which empirically compares the effectiveness of each method. To organize the past literature, we group the methods into three categories: adversarial methods, substitution methods, and LLM methods.

\subsection{Adversarial methods}

Adversarial attack methods are designed to test the adversarial robustness of classifiers by making small, imperceptible changes to the input such as changing a single character or word to induce a model to make incorrect predictions \cite{goodfellow2015explainingharnessingadversarialexamples}. In contrast, counterfactual explanation methods are designed to produce noticeable, realistic, and understandable changes in order to provide useful explanations of a model's behavior.

While adversarial methods were not originally designed to generate counterfactuals, these methods can be repurposed to generate them due to their ability to make minimal input modifications that change model outputs. However, unlike methods for generating counterfactual explanations, adversarial methods generally do not prioritize producing realistic or natural changes and consequently counterfactuals generated using these methods tend to score poorly on metrics measuring plausibility. Some examples of adversarial attack methods include HotFlip (2018) \cite{ebrahimi2018hotflipwhiteboxadversarialexamples} and TextFooler (2020) \cite{jin2020bertreallyrobuststrongtextfooler}.

\textbf{HotFlip.} Introduced by Ebrahimi et al. in 2018, HotFlip \cite{ebrahimi2018hotflipwhiteboxadversarialexamples} is a gradient-based adversarial method which generates adversarial examples or counterfactuals by replacing one or more tokens in the text. To generate a counterfactual, first several candidate substitutions are generated for each position of the original text. The value of each candidate substitution is determined by calculating the partial derivative of the classifier's output with respect to each candidate substitution token. Using beam search, HotFlip explores the space of possible substitutions, prioritizing those with the highest estimated impact on the model's output. The search process continues until the classifier's output is changed using a sufficient number of substitutions or until all positions have been considered. HotFlip is designed to flip the output of the classifier with as few edits as possible and therefore the resulting counterfactuals tend to score highly on validity and sparsity. However, since HotFlip does not consider the plausibility of its substitutions, the generated counterfactuals may score poorly on measures of plausibility and appear unnatural to human readers.

\subsection{Substitution methods}

Both substitution-based counterfactual methods, such as CLOSS \cite{fernTextCounterfactualsLatent2021a} and adversarial methods such as HotFlip \cite{ebrahimi2018hotflipwhiteboxadversarialexamples} identify important words in the original sentence in order to minimize the number of substitutions needed to change the output of the classifier. However, unlike adversarial methods, substitution-based counterfactual methods aim to produce realistic counterfactuals by replacing important tokens with plausible alternatives rather than simply choosing the substitutions that have the greatest influence on the classifier. Many early methods for generating counterfactuals for text classifiers, including REP-SCD (2020) \cite{yangGeneratingPlausibleCounterfactual2020a}, MiCE (2021) \cite{ross2021mice}, and CLOSS (2021) \cite{fernTextCounterfactualsLatent2021a}, fall into this category.

\textbf{CLOSS.} Developed by Pope and Fern \cite{fernTextCounterfactualsLatent2021a} in 2021, CLOSS (Counterfactuals via Latent Optimization and Shapley-guided Search) is a method for generating counterfactual explanations that uses plausible substitutions to find realistic counterfactuals. Similar to HotFlip, CLOSS uses beam search to search through the space of possible substitutions for a valid counterfactual. However, unlike HotFlip, which considers all words in the vocabularly as possible substitutes, CLOSS first uses a masked language model (MLM) such as BERT to select the most plausible substitutes for each position. The value of each substitute is then calculated using partial derivatives similar to HotFlip. CLOSS also uses Shapley values to calculate the value of each substitution which involves calculating the marginal value of a substitution when it is added to an existing set of substitutions. One variant of CLOSS also optimizes the embedding of the input to flip the classifier's output before passing this embedding through the language model to produce candidate substitutes. They evaluate their method on the IMDB sentiment analysis and QNLI datasets according to four metrics: failure rate (validity), percentage of tokens changed (sparsity), BLEU (sparsity), and perplexity (plausibility). They find that CLOSS produces more plausible counterfactuals than methods from the adversarial robustness literature such as HotFlip and TextFooler.

\begin{table*}[ht]
  \caption{Comparison of CLOSS variants.}
  \label{tab:closs-variants}
  \begin{tabularx}{\textwidth}{>{\centering\arraybackslash}X
                              >{\centering\arraybackslash}X
                              >{\centering\arraybackslash}X
                              >{\centering\arraybackslash}X
                              X}
    \toprule
    Method & Optimization & LM head retraining & Shapley values & Description \\
    \midrule
    CLOSS & Yes & Yes & Yes & The original unablated CLOSS variant. \\
    CLOSS-EO & No & No & Yes & More failures but lower perplexity. \\
    CLOSS-RTL & Yes & No & Yes & No change in perplexity but a higher failure rate. \\
    CLOSS-SV & Yes & Yes & No & Removing Shapley values causes a degradation in performance across all metrics. \\
    \bottomrule
  \end{tabularx}
\end{table*}

\subsection{LLM methods}

Recent advancements in large language models (LLMs) have led to the development of new methods for generating counterfactual explanations for text classifiers. While several substitution-based methods such as CLOSS use LLMs to generate plausible word substitutions, these approaches often involve complex, task-specific architectures that limit their generalizability. Additionally, since LLMs have a limited role in the generation of counterfactuals, these methods fail to fully leverage the potential of LLMs for generating high-quality, plausible text counterfactuals.

Recently, several methods for generating text counterfactuals have been developed that leverage LLMs more extensively in the process of generating counterfactuals. These methods can be classified into two categories: controlled generation and general-purpose LLM methods. Early controlled generation methods such as Polyjuice \cite{wuPolyjuiceGeneratingCounterfactuals2021a} and GYC \cite{madaanGenerateYourCounterfactuals2021a} use fine-tuned LLMs such as GPT-2 \cite{Radford2019LanguageMA} to generate counterfactuals. In contrast, general-purpose LLM methods such as FIZLE \cite{bhattacharjeeZeroshotLLMguidedCounterfactual2024} use a custom system prompt to generate counterfactuals, eliminating the need for task-specific fine-tuning.

\textbf{Polyjuice.} Wu et al. (2021) \cite{wuPolyjuiceGeneratingCounterfactuals2021a} developed this controlled generation method which is powered by a GPT-2 model fine-tuned on sentence and counterfactual pairs. The model is fine-tuned on triplets with the form $(x, c, \hat{x})$ where $x$ is the original sentence, $\hat{x}$ is its counterfactual sentence, and $c$ is a control code calculated using a part-of-speech tagger. A counterfactual can then be generated from the fine-tuned model by inputting a sentence $x$ and a control code $c$ such as 'negation' or the control code can be omitted and it will be automatically generated by the model.

\textbf{FIZLE}. Framework for Instructed Zero-shot Counterfactual Generation with LanguagE Models (FIZLE) is a recent method introduced by Bhattacharjee et al. (2024) \cite{bhattacharjeeZeroshotLLMguidedCounterfactual2024} that generates counterfactual text inputs by prompting state-of-the-art LLMs without any task specific fine-tuning or architectures. FIZLE has two variants: The first 'naive' variant simply instructs an LLM model such as GPT-4 to generate counterfactuals via its system message. The second 'guided' method is more complex and involves two stages. First, the model is prompted to identify important input words before being instructed to edit a minimal subset of those words to generate a counterfactual. Results show that FIZLE can generate high-quality counterfactuals and sometimes outperform more specialized methods such as Polyjuice on metrics measuring validity and sparsity.

\section{Experimental setup}

In this section, we describe our experiment to compare the effectiveness of several methods for generating counterfactual explanations for a BERT text classifier. We outline the datasets used, the text classification model to be explained, the counterfactual methods tested, and our evaluation criteria. Our experiment evaluates five methods for generating counterfactual explanations across two datasets using three evaluation metrics \footnote{Source code available at: https://github.com/smcaleese/text-counterfactual-explanation-methods}.

\subsection{Classifier and datasets}

The black-box model to be explained is a BERT-base \cite{devlin2019bertpretrainingdeepbidirectional} binary classification model trained by TextAttack \cite{morris2020textattackframeworkadversarialattacks} and fine-tuned on either the SST-2 or QNLI dataset depending on which dataset is being tested. The model has 12 transformer layers, 110 million parameters, and is uncased. The input to the model can be any text sentence and the output is always a binary label. For example, given a sentence $x$ such as \textit{"I liked the movie"}, the model $f: x \rightarrow y$ classifies the sentence as either 0 for negative sentiment or 1 if positive.

Our two datasets are SST-2 \cite{socher-etal-2013-recursive}, a binary sentiment classification dataset, and QNLI \cite{wang2019gluemultitaskbenchmarkanalysis}, a binary natural language inference dataset. The SST-2 dataset consists of approximately 70,000 short sentences describing movie reviews where each sentence is classified as positive or negative sentiment. The QNLI dataset contains approximately 100,000 rows. Each row consists of a question, a sentence, and a label which is either "entailment" if the sentence contains the answer to the question or "not\_entailment" otherwise. We sample 1000 datapoints from each dataset and pre-process each sentence by converting all characters to lowercase and removing unnecessary spaces.

\subsection{Methods}

We evaluate and compare five methods for generating counterfactual explanations for our text classifier. The following list describes the implementation details and hyperparameters for each method:

\begin{enumerate}
    \item \textbf{HotFlip (2018).} The original implementation of HotFlip makes character-level substitutions. We instead use the implementation of HotFlip from Pope and Fern (2021) \cite{fernTextCounterfactualsLatent2021a} which makes token-level substitutions and is designed for WordPiece classifiers like BERT. We use the hyperparameters: $w = 5$, $b = 15$, $K = 30$, and $t = 0.3$ where $w$ is the number of times each substitution is evaluated using Shapley values, $K$ is the number of candidate substitutions generated for each token index, $b$ is the beam width in the beam search process, and $t$ is the maximum tree depth which prevents no more than 30\% of the tokens in the original text from being modified.
    \item \textbf{CLOSS (2021).} We use the same hyperparameters for CLOSS as HotFlip and set an additional hyperparameter $substitutions\_after\_loc = 0.3$ to ensure that no more than 30\% of tokens can be modified. We use the simpler CLOSS-EO variant of CLOSS which does not require embedding optimization or language model retraining though it includes Shapley value estimations for each candidate substitution.
    \item \textbf{Polyjuice (2020).} We use the original implementation of Polyjuice from Wu et al. (2021) \cite{wuPolyjuiceGeneratingCounterfactuals2021a} For the SST-2 sentiment dataset, we use the "negation" control code and we omit the selection of a control code for the QNLI dataset so that it can be generated automatically.
    \item \textbf{FIZLE-naive (2024).} To implement FIZLE-naive, we used the system prompt from the original paper \cite{bhattacharjeeZeroshotLLMguidedCounterfactual2024} and the GPT-4-Turbo model from OpenAI.
    \item \textbf{FIZLE-guided (2024).} We implement FIZLE-guided using the GPT-4-Turbo model and the two FIZLE-guided system prompts from the original paper \cite{bhattacharjeeZeroshotLLMguidedCounterfactual2024}.
\end{enumerate}

\subsection{Evaluation metrics}

Given an input sentence, each method for generating counterfactual explanations outputs a minimally edited counterfactual sentence that changes the output of the classifier. We use three evaluation metrics to evaluate the average quality of the counterfactuals generated by each method. These metrics are designed to measure the key qualities high-quality counterfactuals should have including high validity, sparsity, and plausibility.

\begin{enumerate}
    \item \textbf{Label flip score (validity).} The label flip score (LFS) measures the mean proportion of counterfactuals which successfully flip the output of the classifier. A high score means that the counterfactual method can reliably generate valid counterfactuals. The label flip score is computed as:
    \begin{equation}
    \text{LFS} = \frac{1}{n} \sum_{i=1}^{n} \mathbbm{1} [f(x_i) \neq f(x'_i)]
    \end{equation}
    \item \textbf{Mean normalized Levenshtein similarity (sparsity).} This metric measures the mean similarity between the original sentences and their counterfactual sentences, normalized to a value between 0 and 1 \cite{levenshtein1966binary}. A higher score indicates higher sparsity, meaning fewer changes were made to generate the counterfactual. The metric is calculated by subtracting the mean normalized Levenshtein distance from 1.
    \item \textbf{Median perplexity (plausibility).} Following Pope and Fern (2021) \cite{fernTextCounterfactualsLatent2021a}, we use the median perplexity, computed as the exponentiated language modeling loss of a GPT-2 model, to assess the linguistic plausibility of the generated counterfactuals. Lower perplexity scores indicate more natural and plausible text. The median is chosen over the mean to mitigate the impact of outliers.
\end{enumerate}

\section{Results}

Our experimental results, summarized in Table 2 and Figure 1, demonstrate the performance of five counterfactual generation methods on the SST-2 and QNLI datasets, with each method evaluated using our three evaluation metrics: label flip score (LFS), mean normalized Levenshtein similarity, and perplexity.

On the SST-2 dataset, CLOSS achieves the highest Label Flip Score (LFS) (0.96), demonstrating its superior ability to generate valid counterfactuals that change the model's prediction. HotFlip achieves an LFS of 0.63 while also having the highest similarity score for the dataset (0.86). FIZLE-naive and FIZLE-guided also achieve high LFS scores (0.88 and 0.85 respectively), which is remarkable given that they are black-box methods without any access to the classifier's internal gradients. While most methods achieve high LFS and similarity scores, Polyjuice fails to achieve a high LFS (0.35) or similarity score (0.53) although it has the ability to generate low-perplexity (139) counterfactuals. Similarly, both FIZLE variants produce highly natural and plausible counterfactual sentences with low perplexity scores (277 and 312). For substitution-based methods, we generally observe a trade-off between sparsity and perplexity: a greater number of substitutions decreases the similarity score while increasing perplexity. Interestingly, this relationship is inverted for LLM methods on the SST-2 dataset. For these methods, lower similarity scores tend to be associated with lower perplexity since the LLMs have the ability to rewrite the original text to have lower perplexity. Therefore, Polyjuice's unusually low perplexity score on the SST-2 dataset can be attributed to its counterfactual sentences having relatively low average similarity to the original sentences.

\begin{table*}[ht]
    \centering
    \small
    \caption{Performance comparison of five counterfactual generation methods on the SST-2 and QNLI datasets. Metrics: Label Flip Score (LFS, higher is better), mean normalized Levenshtein similarity (L.Sim, higher is better), and Perplexity (PPL, lower is better). Best scores for each metric and dataset are in bold.}
    \label{tab:results}
    \begin{tabular*}{\textwidth}{@{\extracolsep{\fill}}lcccccc}
        \toprule
        \multirow{2}{*}{\textbf{Method}} & \multicolumn{3}{c}{\textbf{SST-2}} & \multicolumn{3}{c}{\textbf{QNLI}} \\
        \cmidrule(lr){2-4} \cmidrule(lr){5-7}
        & LFS $\uparrow$ & L.Sim $\uparrow$ & PPL $\downarrow$ & LFS $\uparrow$ & L.Sim $\uparrow$ & PPL $\downarrow$ \\
        \midrule
        HotFlip & 0.63 & \textbf{0.86} & 653 & 0.88 & 0.92 & 132 \\
        CLOSS & \textbf{0.96} & 0.75 & 489 & \textbf{0.99} & \textbf{0.95} & 102 \\
        Polyjuice & 0.35 & 0.53 & \textbf{139} & 0.39 & 0.69 & 80 \\
        FIZLE-naive & 0.88 & 0.70 & 277 & 0.42 & 0.77 & \textbf{69} \\
        FIZLE-guided & 0.85 & 0.72 & 312 & 0.26 & 0.85 & 80 \\
        \bottomrule
    \end{tabular*}
\end{table*}

For the QNLI dataset results, we observe that CLOSS maintains strong performance, achieving the highest LFS (0.99) and similarity score (0.95) on the dataset. HotFlip also demonstrates a high LFS (0.88) and similarity scores (0.92), although it's counterfactuals consistently have higher perplexity (132) compared to CLOSS (102) and the other methods. Although LLM-based methods generate plausible, low-perplexity counterfactuals for this dataset, these methods generally fail to achieve high validity scores. For example, FIZLE-naive achieves a validity score of only 0.42 on this dataset. These results suggest that for some datasets such as QNLI, the importance of each token for the classifier's output cannot always be accurately determined without direct access to the classifier's gradients. Consequently, white-box methods such as CLOSS generate counterfactuals with significantly higher validity than LLM-based methods on this dataset. Counterfactual sentences for the QNLI dataset consist of question-answer pairs separated by a [SEP] token. We found that Polyjuice often produced only questions as outputs for the QNLI dataset. As a result, a formatting function was required to append the [SEP] and answer tokens to Polyjuice outputs in order to make them valid.

\begin{figure}
    \centering
    \includegraphics[width=12cm]{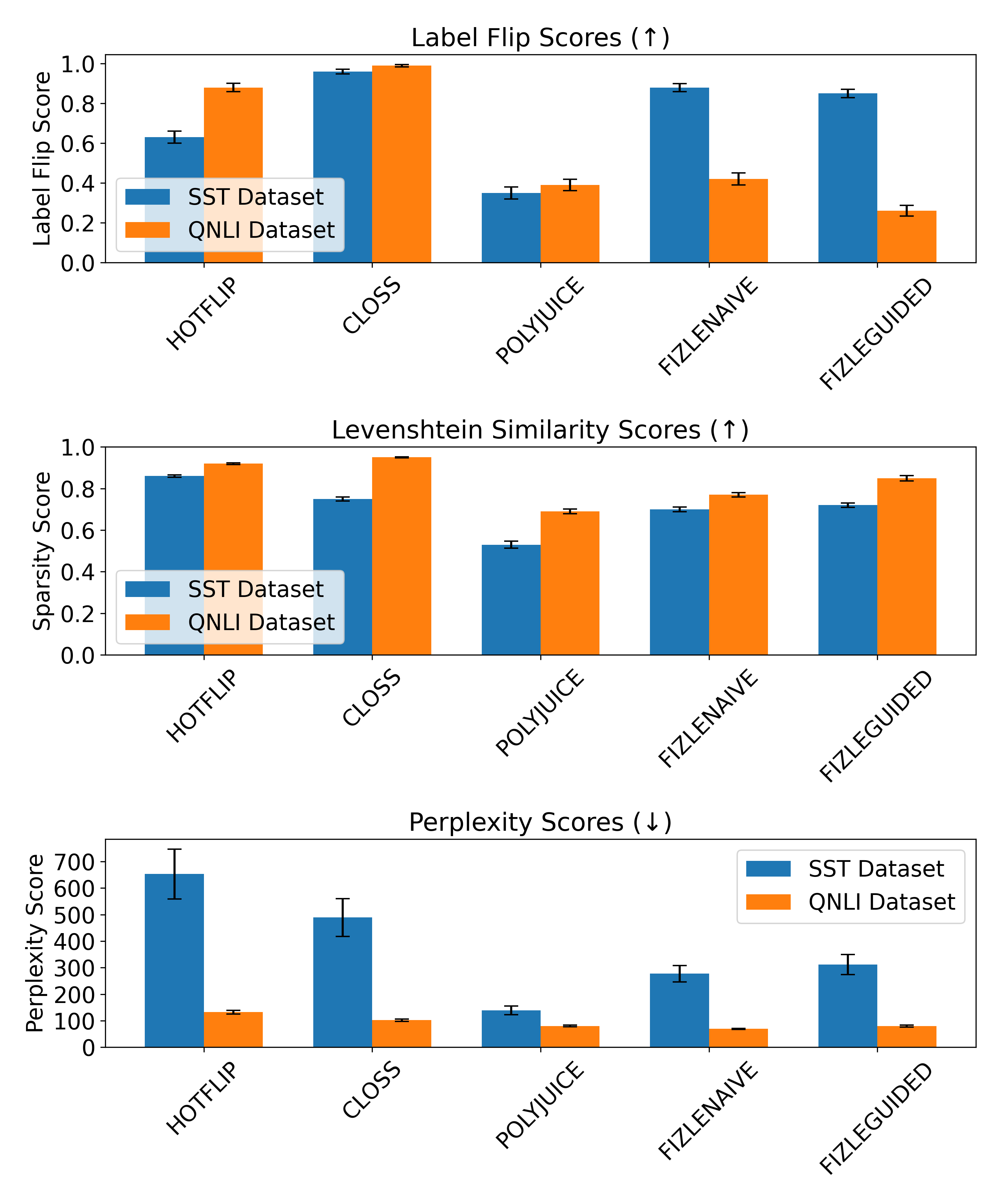}
    \caption{Figure illustrating the results of our comparative analysis. Error bars are bootstrapped 95\% confidence intervals.}
    \label{fig:results-figure}
\end{figure}

\section{Conclusion}

In this work, we evaluated five methods for generating counterfactual explanations across two datasets, evaluated using three metrics. We aimed to address two primary research questions: 1) Which methods for generating counterfactual explanations for text classifiers are most effective? 2) Do newer, simpler methods that prompt general-purpose LLMs to generate counterfactuals outperform older, more specialized methods?

Our findings suggest that the effectiveness of each counterfactual method varies across evaluation metrics and datasets. For example, both FIZLE-naive variants achieve high validity, sparsity, and plausibility scores on the SST-2 dataset but fail to maintain high validity scores on the QNLI dataset. Conversely, CLOSS consistently achieves high validity and similarity scores but underperforms LLM-based methods on measures of plausibility. These results suggest that the most effective method depends on which dataset is used and which evaluation metric is prioritized.

Regarding the second question, we find that newer LLM-based methods such as FIZLE and Polyjuice consistently produce more plausible and natural counterfactuals than substitution-based methods. However, these methods often fail to produce valid counterfactuals, especially on some datasets such as QNLI. These results indicate that newer LLM-methods outperform older methods on some metrics but not others.

Our results reveal a performance trade-off for text counterfactual explanation methods. White-box methods like CLOSS, which have access to the classifier's gradients, consistently produce valid counterfactuals but may sacrifice some linguistic plausibility. Conversely, black-box LLM methods generate more natural and plausible text but do not always succeed in generating valid counterfactuals that change the classifier's output.

For future work, we suggest developing methods for generating counterfactual explanations that achieve high validity, sparsity and plausibility by combining the strengths of gradient-based and LLM-based methods. In conclusion, this work contributes to the growing body of research on explainable AI in NLP and counterfactual explanations and the broader goal of ensuring that NLP models are understandable and trustworthy.

\bibliography{references}

\end{document}